\documentclass[conference]{IEEEtran}
\IEEEoverridecommandlockouts
\usepackage{amsmath,amssymb,amsfonts}
\usepackage{algorithmic}
\usepackage{graphicx}
\usepackage{textcomp}
\usepackage{xcolor}
\def\BibTeX{{\rm B\kern-.05em{\sc i\kern-.025em b}\kern-.08em
    T\kern-.1667em\lower.7ex\hbox{E}\kern-.125emX}}

\usepackage[backend=biber]{biblatex}
\usepackage[ruled,vlined]{algorithm2e}
\usepackage[english]{babel}
\usepackage{csquotes}
\usepackage{subfig}

\usepackage{multirow}
\usepackage{colortbl}
\usepackage{hhline}
\usepackage{booktabs}
\usepackage{textgreek}
\usepackage{threeparttable}
\usepackage{mathtools}
\usepackage{lipsum}
\usepackage{multicol}

\usepackage{cuted}
\usepackage{amsmath}
\usepackage[font=small]{caption}
\usepackage{bigints}

\usepackage[ruled,vlined]{algorithm2e}
\SetAlFnt{\scriptsize}

\usepackage{amsthm}
\usepackage{amssymb}

\usepackage{tabularx}

\usepackage{tabu}

\newcolumntype{M}[1]{>{\centering\arraybackslash}m{#1}}

\usepackage{hyperref}
\hypersetup{
    colorlinks=true,                
    breaklinks=true,                
    urlcolor= black,                
    linkcolor= blue,                
    bookmarksopen=false,
    filecolor=black,
    citecolor=blue,
    linkbordercolor=blue
}

\usepackage{booktabs}

\graphicspath{ {./figs/} }
\usepackage{amsmath}
\DeclareMathOperator*{\argmin}{arg\,min}

\DeclarePairedDelimiter\abs{\lvert}{\rvert}%
\makeatletter
\let\oldabs\abs
\def\abs{\@ifstar{\oldabs}{\oldabs*}}
\let\oldnorm\norm
\def\norm{\@ifstar{\oldnorm}{\oldnorm*}}
\makeatother

\begin{document}

\title{Local Trajectory Planning For UAV Autonomous Landing}

\author{Yossi Magrisso, Ehud Rivlin and Hector Rotstein


\thanks{All authors are with \textit{The Henry and Marilyn Taub Faculty of Computer Sciences, Technion - Israel Institute of Technology, Haifa, Israel}. \newline
\tt\small{yossimag@alumni.technion.ac.il,  \newline ehudr@cs.technion.ac.il,  \newline hector@ee.technion.ac.il } 
 
} 

}

\maketitle

\begin{abstract}

An important capability of autonomous Unmanned Aerial Vehicles (UAVs) is autonomous landing while avoiding collision with obstacles in the process. Such capability requires real-time local trajectory planning. Although trajectory-planning methods have been introduced for cases such as emergency landing, they have not been evaluated in real-life scenarios where only the surface of obstacles can be sensed and detected. We propose a novel optimization framework using a pre-planned global path and a priority map of the landing area. Several trajectory planning algorithms were implemented and evaluated in a simulator that includes a 3D urban environment, LiDAR-based obstacle-surface sensing and UAV guidance and dynamics. We show that using our proposed optimization criterion can successfully improve the landing-mission success probability while avoiding collisions with obstacles in real-time.      

\end{abstract}


\section{Introduction}

Providing autonomous features to UAVs is of major importance. Autonomy can free human operators from flight tasks, reduces the dependency on wireless communication, and also increases the number of platforms operating safetly and simultaneously in a given area. Among other important autonomous features, navigation, obstacle collision avoidance and landing sites exploration are of interest in the present work. Having these features can enable solving the ''last mile" problem for package delivery or for applications requiring provisions for  emergency landing \cite{hedayatpour2018path}.

When performing an autonomous mission, a UAV will usually execute a global path, pre-calculated to be optimal for some significant performance criterion. As opposed to this, a \emph{Local} Trajectory planner (LTP) has the objective of calculating and executing a local path that attempts to follow the global path as close as possible while avoiding collision with obstacles that were not considered in the pre-computations \cite{gao2020teach}. A local Planner has to be efficient since it needs to run in soft real-time and possibly recalculate the trajectory whenever a new obstacle is detected. However, optimizing  the trajectory to some extent is most desirable.

A risk-based trajectory planning approach was introduced in \cite{9213982,primatesta2019risk}. In this work a risk map is used as an input for the optimization process, and a modification of the optimal $A^*$ algorithm \cite{4082128}, namely $riskA^*$, is used for planning a trajectory that minimizes the risk when flying over populated areas. This approach can be adopted for the autonomous landing site search, with a difference of using an \textit{opportunity} map instead of a risk map. 

For obstacle collision avoidance LTP algorithms usually require minimal a-priori information of the environment since sensor-based data is more relevant and important. For instance, the bio-inspired Bug family of algorithms \cite{MCGUIRE2019103261,doi:10.1177/027836499801700903}  plan local trajectory in the continuous configuration space by following a direct path to the goal point while bypassing obstacles along their surface border whenever they are encountered. More aggressive success-oriented methods attempt to find local trajectories by using optimized Hermite polynomials \cite{LinKong} or Bézier curves \cite{hu_li_he_han_2019,Elmokadem1,Cimurs,ingersoll2016uav,gao2020teach}. The optimization criteria can include minimizing length, curvature, time-of-flight, or energy consumption, or maximizing smoothness and consistency. Although these algorithms work well in some cases, they are neither optimal nor complete.   

Realistic obstacle detection and mapping must be taken into account when evaluating performance of trajectory-planning algorithms. Most sensors can sense and map only the outer surfaces of obstacles but not their interior volume, and this can have a major impact on the feasibility and quality of the planned trajectories. 

We aim to find efficient LTP algorithms that can both help avoid obstacle collision in realistic sensor-based obstacle detection scenarios, \textit{and} optimize the trajectory for the landing site search task. In \ref{ch:ProblemFormulation} we formulate the problem and optimization criteria, in \ref{CH:LTP} we describe the algorithms that were evaluated, and in \ref{Simulation_results}  evaluation results are presented.


\section{Problem Formulation}
\label{ch:ProblemFormulation}
\subsection{Landing site search process}

\begin{figure}
    \centering
    \includegraphics[width=0.5\textwidth]{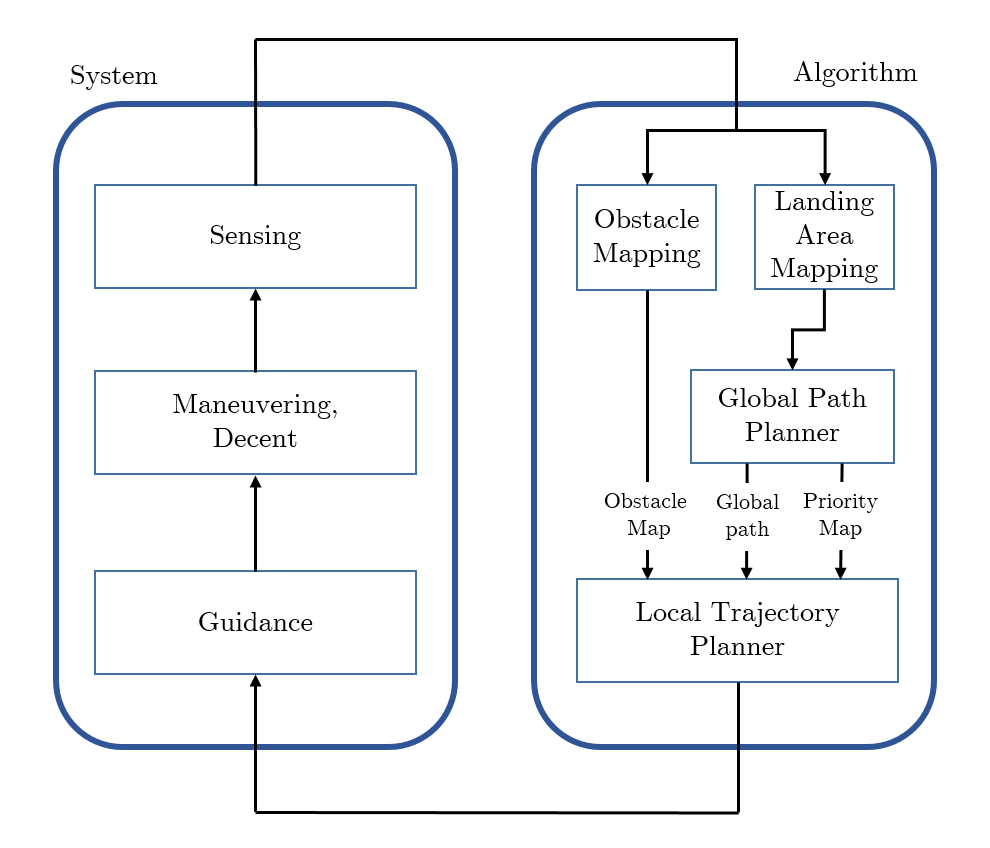}
    \caption{System diagram of the autonomous landing process.}
    \label{System_diagram}
\end{figure}

The main motivation for this paper is the probabilistic multi-resolution approach for finding a suitable landing spot introduced by Pinkovich et al. in \cite{BarakP_MultiRes}. In that work, a search is conducted by calculating a feasible global flight path at a given altitude using prior information. The global path is such that a look-down camera installed on the UAV captures images from areas with high a-priori probability of being suitable for landing. An ''Oracle" module then uses the images to calculate a post-priory probability map of the imaged area. Clearly, this requires that the flight path be executed safely by avoiding the collision with obstacles. Once probabilities have been updated, the UAV is lowered to a new altitude and a global path is recomputed. The process continues iteratively until a pre-specified threshold of suitability for landing has been exceeded, or else failure is declared. Fig. \ref{System_diagram} demonstrates this iterative process. 

For simplicity, we assume the UAV can be considered to be a 3D point (meaning the configuration space is identical to the work space), and that it has no kinematic constraints. The flight path execution is computed by the LTP using two main outputs from the Global Path Planner (GPP): 1) A set of 3D points (or \emph{way-points}) $\{\vec{r}_{G,k}\}$ specifying the desired global path, and 2) a 2D priority map $p_{l}(x,y)$ prioritizing flight over the different areas on ground (see Fig. \ref{50_GPs_on_DSM}). The higher the priority of a point $(x,y)$, the more likely it is for the look-down camera to take an image of a good landing spot beneath that point. The LTP has to optimize the actual flight trajectory using three optimization criteria: 1) The trajectory has to be as close as possible to the global path; 2) Maximize the average priority along the trajectory, and 3) Obstacle collision must be avoided. At each time instant, an initial point $\vec{r}_{init}(t)$ and a goal point $\vec{r}_{goal}(t)$ are defined for the LTP. Note that $\vec{r}_{init}$ can be either the current or a previous position of the UAV, and $\vec{r}_{goal}=\vec{r}_{G,k}$ is the next global path point to be reached. If $\vec{r}_{G,k}$ is reached or found to be unreachable, it is removed from the way-point list and a local trajectory is planned to the next global point $\vec{r}_{goal}=\vec{r}_{G,k+1}$.

\subsection{Obstacle detection and mapping}

The UAV is assumed to be equipped with at least one or more sensors designed for the specific task of obstacle detection. Such sensors can be LiDAR, Radar, depth camera, or regular front-looking camera.  The detection data of each one of the sensors is given at its own local coordinate system, usually polar and centered around its installation point and direction. For enabling obstacle global mapping, detection coordinates are transformed to a global coordinate system that coincide with the one used for navigation. Navigation data including UAV 3D position and orientation is assume to be perfect.

\begin{figure}
    \centering
    \subfloat[]{\includegraphics[width=0.23\textwidth]{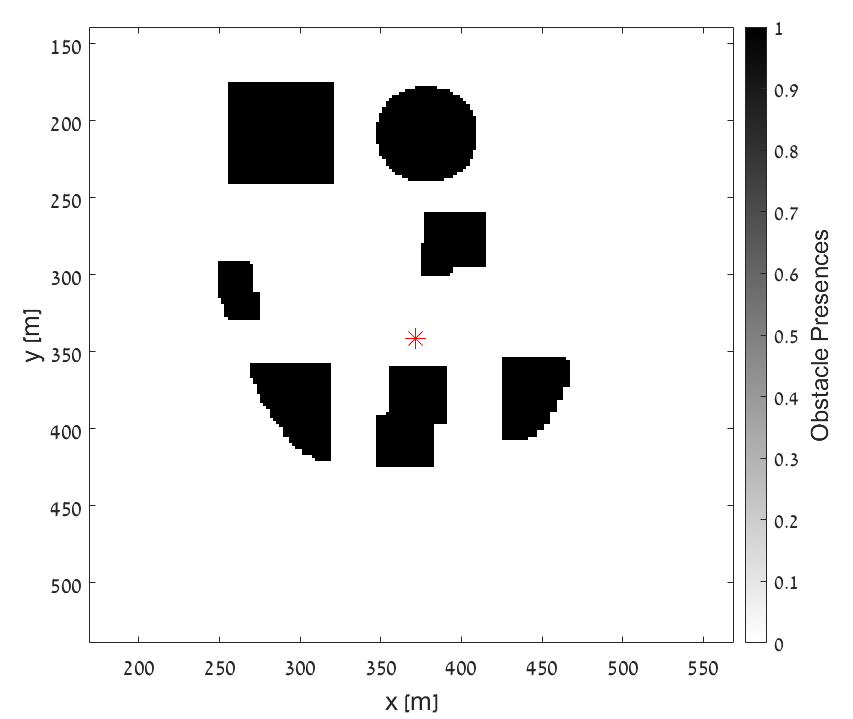}}
    \subfloat[]{\includegraphics[width=0.23\textwidth]{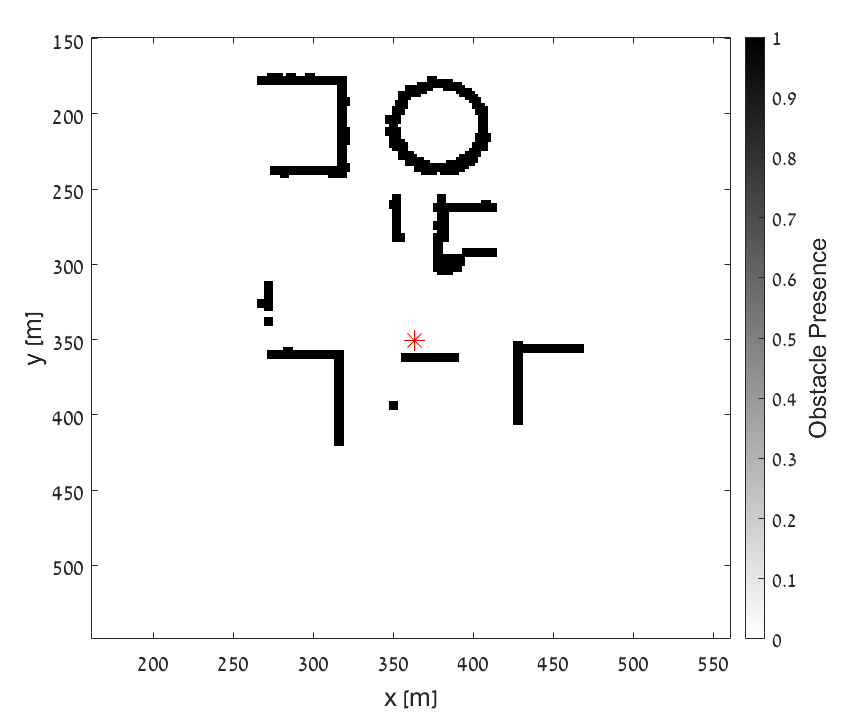}}
    \caption{ The obstacle map is always kept centered around the position of the UAV, indicating obstacle presence. A slice of the 3D map is shown for height z = 50$m$. The maximal detection range was set to be 100$m$ in this case. (a) Volume sensing. (b) Surface sensing. }
    \label{OccMap}
\end{figure}

The information regarding location of obstacles is concentrated in a 3D obstacle map (see Fig.~\ref{OccMap}).  Because of memory consumption and run-time performance considerations, each cell in the obstacle map must have a limited resolution in every axis, and the whole map coverage must also be limited. Due to the coverage limitation, the map is kept centered around the position of the aircraft and shifted constantly for that purpose. Once the obstacle map is updated with the most current locations of obstacles, it  serves as an input for trajectory planning. 

For future reference, let $\mathcal{C}_\text{free}$ and $\mathcal{C}_{\text{obs}}$ be the subsets free and occupied by mapped obstacles, respectively. Let also $\mathcal{C}_\text{reach}$ be the reachable space, meaning the set of all points that can be connected to $\vec{r}_{goal}$ by a trajectory $\vec{\ell}\in\mathcal{C}_\text{free}$ of finite length. 

\subsection{Surface sensing and hollow obstacles}

When sensors detect an obstacle they usually can only sense the outer surface of the obstacle but not its inner volume. The inner volume of the obstacle can never be reached by the sensors, so unless prior knowledge of its existence is given it is always falsely mapped to $\mathcal{C}_\text{free}$ by default. This surface sensing challenges the LTP, especially in cases when a global path point falls inside an obstacle. In these cases only trajectory planning failure can indicate a problem in the global path point (see Fig. \ref{GSWP_path_surface}). The local trajectory planning algorithm must therefore be complete so to reject invalid global path points, but avoid rejecting valid global path points that are in $\mathcal{C}_\text{reach}$.

\subsection{Optimization criteria}

For optimizing the local trajectory $\ell$ between $\vec{r}_{init}$ and $\vec{r}_{goal}$ we define a penalty function $\varepsilon_p(\ell)$ to be minimized that has three main component:

\begin{enumerate}
    \item \textbf{Obstacle avoidance - } The penalty for passing through an obstacle must be very high:
\begin{equation}
\varepsilon_{p0}(\ell) = 
\begin{cases}
     \infty,&\ell \text{ crosses an obstacle}  \\
     0 ,&\text{otherwise}
\end{cases}
\end{equation}

\item \textbf{Minimal trajectory length - } The trajectory should to be as short as possible for saving energy and time. The shortest path between $\vec{r}_{init}$ and $\vec{r}_{goal}$ is $L_{direct}=\|\vec{r}_{goal} -\vec{r}_{init}\|$. A suitable penalty function is a one proportional to the trajectory length $L$:
    \begin{equation}
        L=\int\limits_{\vec{r}_{init}}^{\vec{r}_{goal}}{d\ell}
    \end{equation}\\
     To avoid dependency on the absolute trajectory length, a proper penalty factor is therefore: 
    \begin{equation}
        \varepsilon_{p1} = \dfrac{L}
        {L_{direct}} 
    \end{equation}
    Notice that $L$ is always longer then $L_{direct}$, so the minimal penalty $\varepsilon_{p1}$ will be 1 for $L = L_{direct}$.
\newline
    \item \textbf{Prioritizing flight over certain areas - } Using the priority map given by the global path planner, $0\leq p_{l}(x,y)\leq 1$, an additional term is added to the penalty function that reflects the average priority of regions covered beneath the flight trajectory:
    
\begin{equation}
        \varepsilon_{p2} = -\dfrac{\displaystyle\int_{\ell}{p_{l}(x,y)d\ell}}
        {L} 
\end{equation}\\
The higher the average priority on the trajectory, the lower will be the penalty. Notice that $-1\leq \varepsilon_{p2}\leq 0$, with minimal penalty of $\varepsilon_{p2}=-1$ when $p_l(x_{\ell},y_{\ell})=1$ for all points on the trajectory.

\end{enumerate}

The total penalty is set as a weighed combination of the three penalty functions:

\begin{equation}
        \varepsilon_{p}(\ell) = \varepsilon_{p0}(\ell)+ W_L\cdot \varepsilon_{p1}+W_p\cdot \varepsilon_{p2}
\label{optim}
\end{equation}

with $W_L$, $W_p$ being some weighting constants that reflect the preference between the different optimization criteria. This optimization criterion resembles the risk function in \cite{primatesta2019risk}, but instead of risk it takes into account look-down camera coverage for landing site search.
$\varepsilon_{p1}$ and $\varepsilon_{p2}$ can alternatively be set as two separate cost function for bi-objective path planning algorithms such as BOA* \cite{Hernández_Ulloa_Yeoh_Baier_Zhang_Suazo_Koenig_2020}.

\section{Local trajectory planning} 
\label{CH:LTP}

\subsection{Trajectory planning using B\'ezier curves}

\begin{figure}
    \centering
    \includegraphics[width=0.4\textwidth]{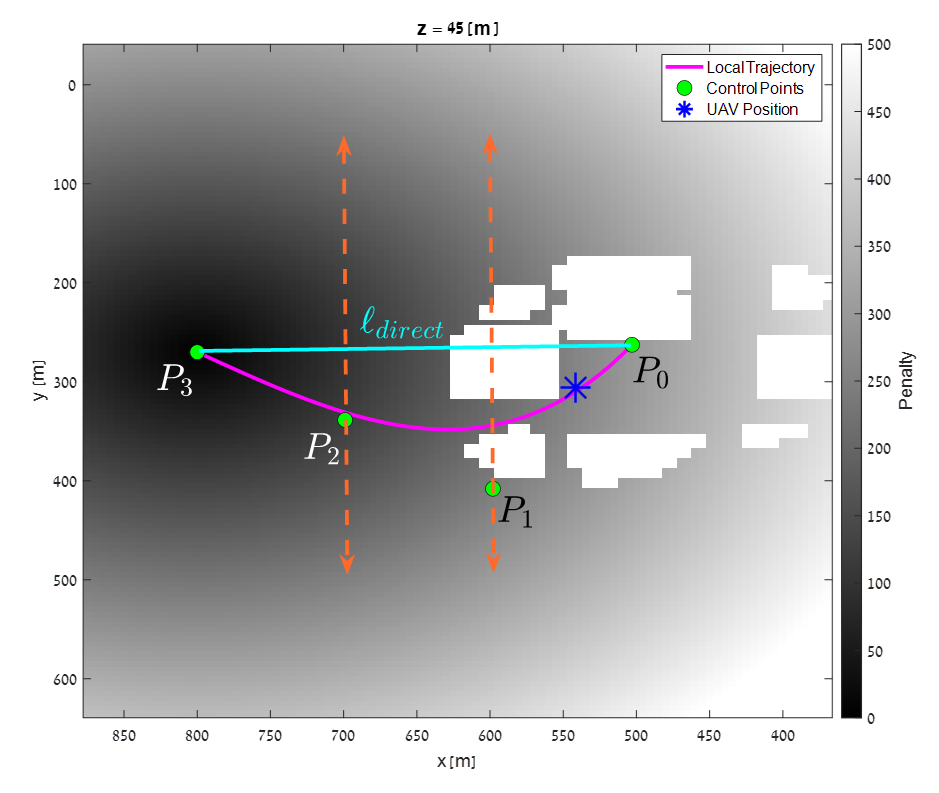}
    \caption{Local trajectory planning using a B\'ezier curve (magenta line) in a volume detection scenario. The B\'ezier control points ($P_n$) are marked in green, and the penalty map is shown at the background. Cells representing  obstacle presence receive a high penalty. An exhaustive search is performed on points $P_1$ and $P_2$, moving them along the dashed orange arrows that are perpendicular to the line $\ell_{direct}$ (in cyan). The optimal local trajectory is the one minimizing the cumulative penalty on the curve.}
    \label{Penalty_map_t_201}
\end{figure}

A possible way of finding an optimal trajectory between $\vec{r}_{init}$ and $\vec{r}_{goal}$ is by using a cubic B\'ezier curve with four control points, $\vec{\ell}\left(s|\{P_n\} \right), n = 0,1,2,3$. $P_n$ are the control points, and $s$ is the parameterization of the curve satisfying $0 \leq s \leq 1$. \small
\begin{equation}
\label{eq:Bezier}
\begin{split}
\vec{\ell}\left(s | \{P_n\} \right) & = 
\left( \left[x_\ell(s),y_\ell(s),z_\ell(s)\right] | \{P_0,P_1,P_2,P_3\} \right) \\
& =  (1-s)^3P_0+3(1-s)^2sP_1+3(1-s)s^2 P_2+s^3 P_3 
\end{split}
\end{equation}
\normalsize
The first and last control points ($P_0$ and $P_3$) are set as $\vec{r}_{init}$ and  $\vec{r}_{goal}$ respectively. For $s=0$ the trajectory begins at $P_0$, and for $s=1$ the trajectory ends at $P_3$. The search focuses on the middle control points, $P_1$ and $P_2$, as to minimize $\varepsilon_p$:\\
\begin{equation}
    \left(P_1^*,P_2^* \right) =
    \argmin_{
                    P_1,P_2 
            }
              \left(\varepsilon_p\right) 
\end{equation}\\
$P_1$ and $P_2$ are initially set on line $\ell_{direct}$ connecting between $P_0$ and $P_3$, spaced evenly between them as shown in Fig.~\ref{Penalty_map_t_201}. An exhaustive search is performed on $P_1$ and $P_2$ by moving both of them on the $xy$ plane, in perpendicular direction to  $\ell_{direct}$. A main assumption is that change in height  (movement in the $z$ axis) is prohibited or is significantly limited to the very least. Nevertheless, to avoid singularity in some scenarios some limited search in the $z$ axis of points $P_1$ and $P_2$ is also possible. If the search grid for $P_1$ and $P_2$ is of size $M$, then the complexity of the algorithm is $\mathcal{O}(M^2)$, and does not depend on resolution and coverage of the obstacle map. 

\subsection{Trajectory planning using Cost Wavefront Propagation}

\begin{figure}
     \centering
     \subfloat[]{\includegraphics[width=0.25\textwidth]{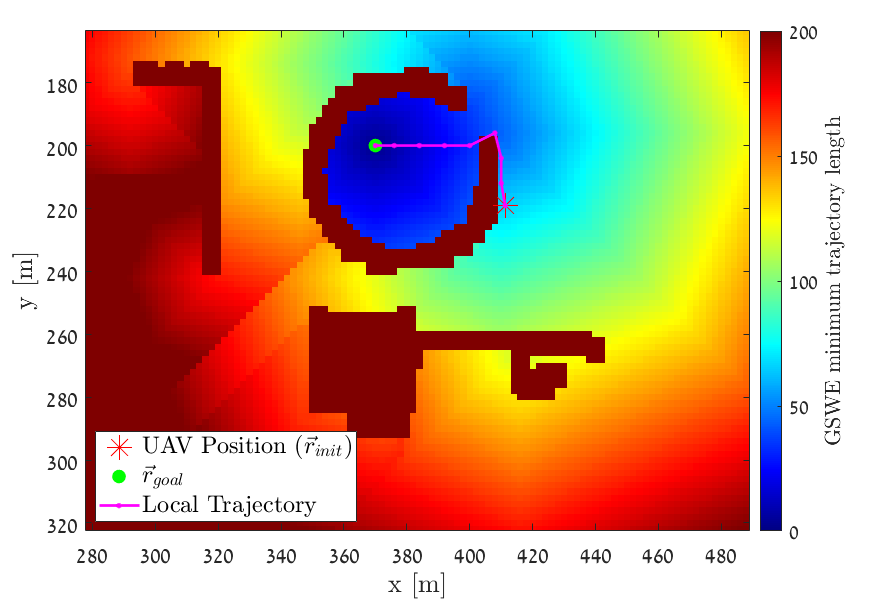}}
      \subfloat[]{\includegraphics[width=0.25\textwidth]{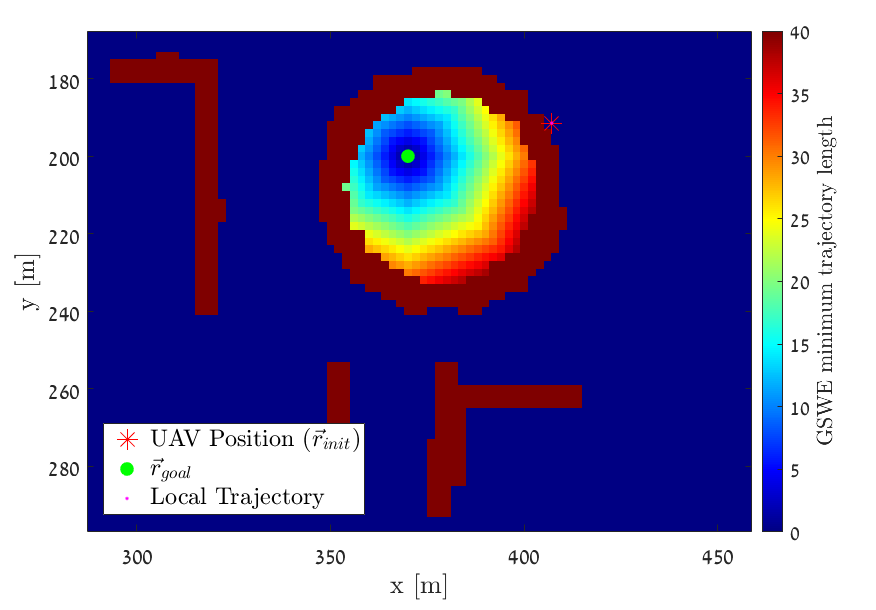}} 
     \caption{Greedy Grid-based Wavefront Propagation (GGWP) trajectory planning in surface detection scenarios where the goal point (green) falls inside an obstacle. (a) The surface of  obstacle is not entirely detected and a path is calculated to the goal point. (b) The surface of the obstacle is entirely mapped and wavefront propagation is limited only to the interior volume of the obstacle. In this case planning failure is declared.}
\label{GSWP_path_surface}
\end{figure}

Finding a local trajectory using a polynomial method such as B\'ezier curves can succeed in simple cases, but can fail in more complicated scenarios such as narrow maze-like corridors or  bug traps. A different solution for these scenarios is therefore required. One of the methods proposed for path planning is by calculating and following a navigation function \cite{Lav06}. A navigation function was generated by a Greedy Grid-based version of the cost Wavefront Propagation algorithm \cite{10.1117/12.949097} (we name \textit{GGWP}). In this algorithm the navigation function construction initiates from the goal point and propagates outward incrementally on the grid until reaching the initial point. At the second stage of the algorithm, after the initial point is reached, the shortest path is determined by following the geodesic line of the navigation function in a gradient decent manor. The GGWP algorithm is able to find trajectories for complicated scenarios such as bug-traps (see Fig. \ref{GSWP_path_surface}) in $\mathcal{O}(N)$ ($N$ being the number of cells in  the occupancy map). However, it aims to only find short trajectories and cannot optimize for the landing site search problem in (\ref{optim}).

\subsection{Trajectory planning using Length-Priority A*}

For minimizing $\varepsilon_{p}$ on the trajectory the well known backwards weighted $A^*$ algorithm \cite{hansen2007anytime,NIPS2003_ee8fe909} was used. A graph is constructed by connecting all neighbouring samples in the regular grid of the obstacle map, and a trajectory search is performed on the graph edges. Similar to $riskA^*$ \cite{9213982}, we also modify the cost-to-go function $g(\vec{r}_i,\vec{r}_{goal})$ and heuristic estimated cost-to-come function $h(\vec{r}_{init},\vec{r}_i)$ of $A^*$, but in a way that supports the length-priority optimization of the trajectory:
\begin{equation}
        g(\vec{r}_i,\vec{r}_{goal}) =  \varepsilon_p(\ell:\vec{r}_i\rightarrow\vec{r}_{goal}) 
\label{A_star_new_g}
\end{equation}
\begin{equation}
        h(\vec{r}_{init},\vec{r}_i)= \dfrac{\|\vec{r}_{init}-\vec{r}_i\|}{L_{direct}}
\label{A_star_new_h}
\end{equation}
\begin{equation}
        f(\vec{r}_i,\vec{r}_{goal}) = g(\vec{r}_i,\vec{r}_{goal}) + W_{A^*} \cdot  h(\vec{r}_{init},\vec{r}_i) 
\label{A_star_new_f}
\end{equation}

with $W_{A^*}$ the weighting factor of the heuristic function, and $f(\vec{r}_i,\vec{r}_{goal})$ the estimated total cost function. We name this form of weighted $A^*$ algorithm \textit{Weighted Length-Priority A*}, or \textit{WLP-$A^*$}. When setting $W_{A^*}=1$ the algorithm becomes a form of $A^*$ we name \textit{Length-Priority A*}, or  \textit{LP-$A^*$}. When setting $W_{A^*}=0$ the algorithm becomes a form of the backwards Dijkstra algorithm we name \textit{Length-Priority Dijkstra}, or \textit{LP-Dijkstra}.

\section{Results} \label{Simulation_results}

\begin{figure}[ht]
    \centering
    \subfloat[]{\includegraphics[width=0.17\textwidth]{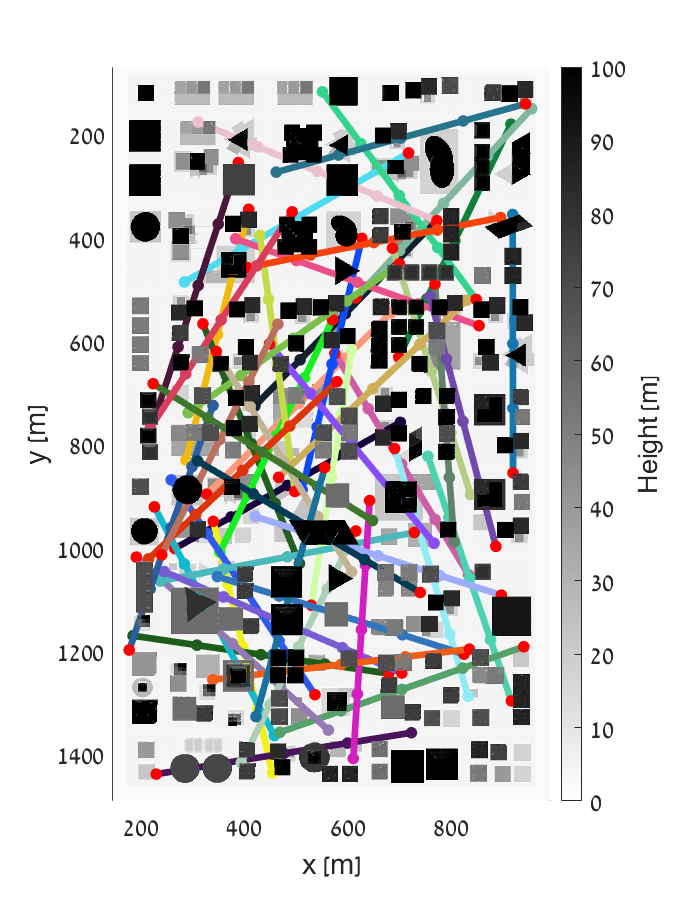}}
    \subfloat[]{\includegraphics[width=0.17\textwidth]{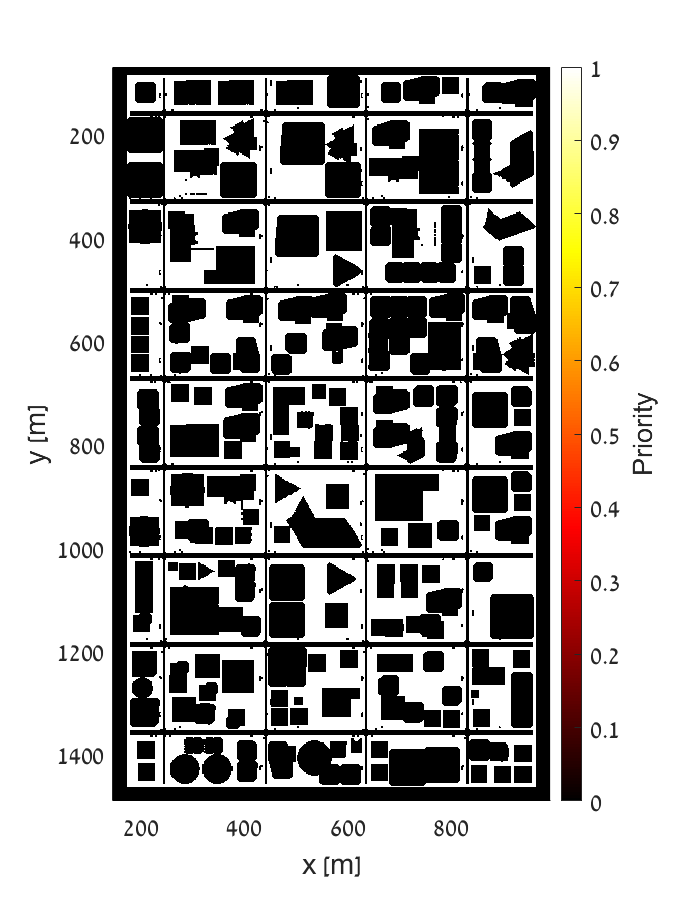}}
     \subfloat[]{\includegraphics[width=0.17\textwidth]{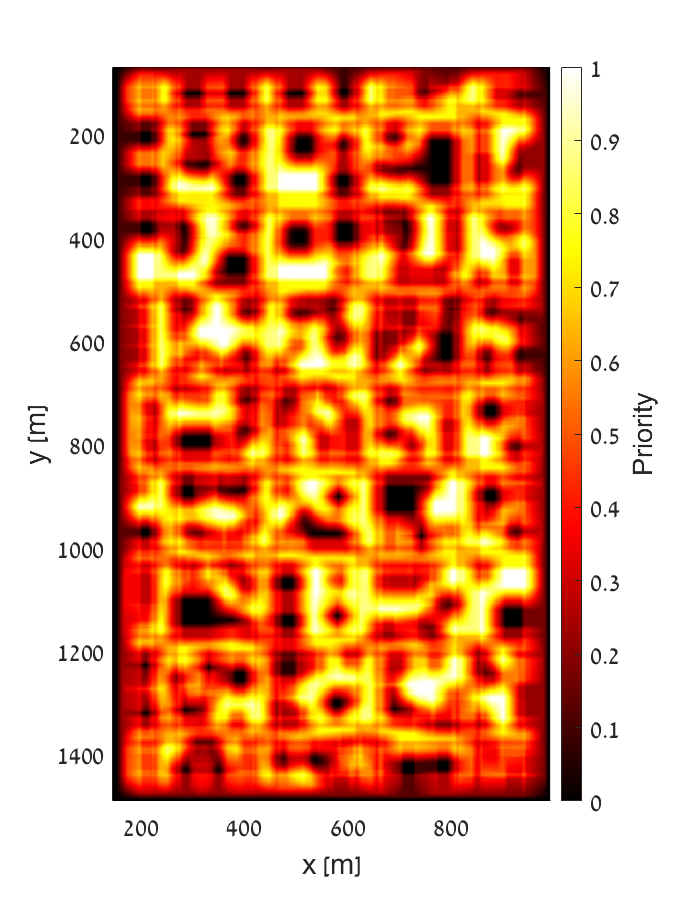}}
    \caption{Statistical analysis for two different priority maps and 50 global paths. In (a) 50 linear global paths chosen randomly are placed on top of the DSM. All of the global paths were set at 50 $m$ height. The red dots mark the starting points of each path. In  (b) an (c) two different priority maps used for the analysis. Map (b) is \textbf{binaric} having the value 1 above valid landing sites. In map (c) low-pass filtering (\textbf{LPF}) was applied for better representing  the down-looking camera coverage beneath each location.}
    \label{50_GPs_on_DSM}
\end{figure}

\begin{figure}[h]
    \centering
    \subfloat[]{\includegraphics[width=0.25\textwidth]{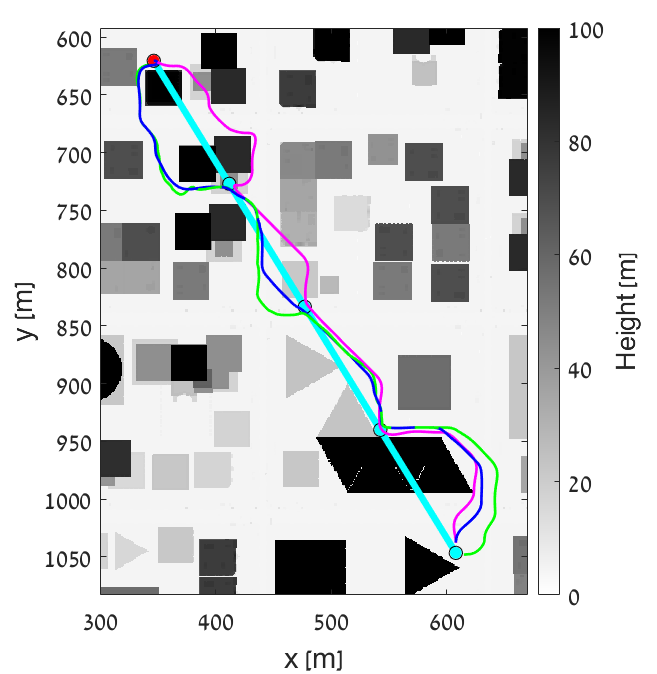}}
    \subfloat[]{\includegraphics[width=0.25\textwidth]{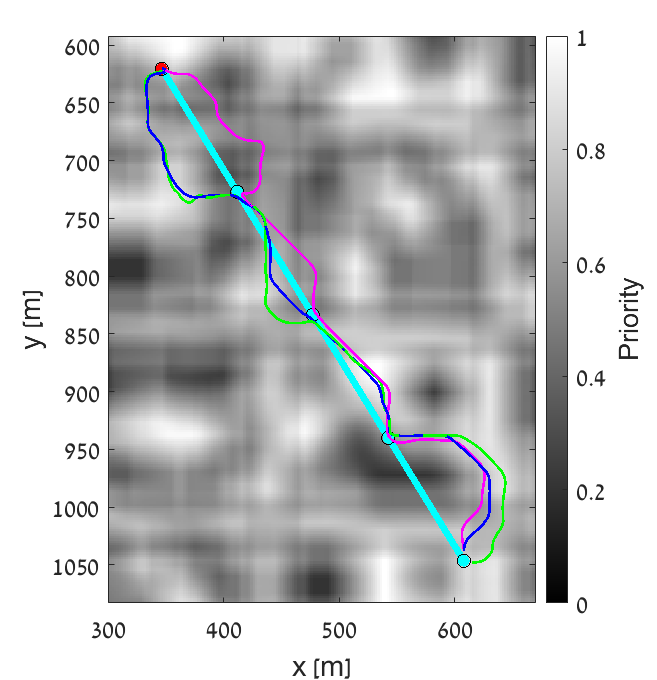}}
    \caption{Local trajectory planning and execution using LP-Dijkstra in an urban environment. The different trajectories are laid over a DSM (a) and over a LPF priority map (b). The global path is marked in dotted cyan line, running at a height of 50$m$, beginning from the top left corner (red point). Three planned trajectories  are shown for different penalty functions:  $W_P/W_L=0$ (magenta),   $W_P/W_L=2$ (blue),  $W_P/W_L=5$ (green). }
    \label{Sim_results1}
\end{figure}

A simulation of an autonomous flying UAV was implemented using Matlab. AirSim simulator \cite{shah2018airsim} was used both for creating a Digital Surface Model (DSM) data of an urban area, for simulating obstacles and their detections, and for generating segmentation masks of elements on the ground and thus priority maps. The priority maps were calculated in two ways: 1. A \textit{binari} map was created by setting a constant value 1 in valid landing regions and 0 in invalid ones; 2. By running a Low-Pass Filter (\textit{LPF}) on the segmentation mask, with a window size similar to the look-down camera footprint. The footprint size is given by  $2h\cdot \text{tan}(FOV/2)$, where the camera Field-Of-View (FOV) used was $53^{\circ}\times53^{\circ}$ and $h$ is the UAV height changing dynamically. For $50m$ height the camera footprint is approximately $50m\times50m$.  For testing the capability of  obstacle avoidance, an artificial global path was used running intentionally through buildings.  A statistical analysis was performed on such 50 randomly chosen linear global paths, all having length of 500$m$. The designated flight velocity was 3 $m/sec$. Different weights $W_L,W_P$ were used for the penalty function. Obstacle detection by surface sensing of LiDAR rays was simulated at maximal range of 50$m$. The obstacle map grid $(x,y,z)$ was of size (200,200,40) ($N=1.6\cdot10^6$ cells) and of resolution (2$m$,2$m$,4$m$) (total map coverage of (400$m$,400$m$,160$m$)). 
The different priority maps and global paths used for the analysis are shown in In Fig \ref{50_GPs_on_DSM}, and in Fig. \ref{Sim_results1} an example is shown for the final trajectories, laid over the DSM and over the priority map.

The parameters for evaluation were the average priority on the trajectory (\textbf{$\overline{P}$}), the average length ratio between the trajectory and the global path (\textbf{$\overline{\frac{L_{LP}}{L_{GP}}}$}), and algorithm run-time. Run-time testing was performed on an Intel\textregistered\,   Core\texttrademark\, i7-7820HQ CPU @ 2.90GHz machine with 16 GB RAM.  
Four algorithms were tested and compared: LP-Dijkstra, LP-$A^*$, WLP-$A^*$ with $W_{A^*}=\sqrt{3}$ and GGWP. The reason for choosing $W_{A^*}=\sqrt{3}$ was for better comparison of WLP-$A^*$ with GGWP, since both their path lengths are upper bounded by a $\sqrt{3}$ factor from the shortest path length on the grid.

\begin{figure}
    \centering
     \subfloat[]{\includegraphics[width=0.5\textwidth]{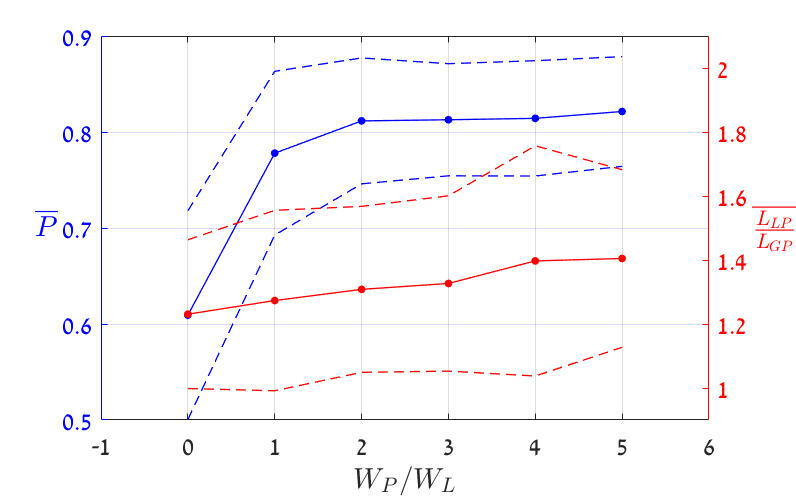}}\\
     \subfloat[]{\includegraphics[width=0.5\textwidth]{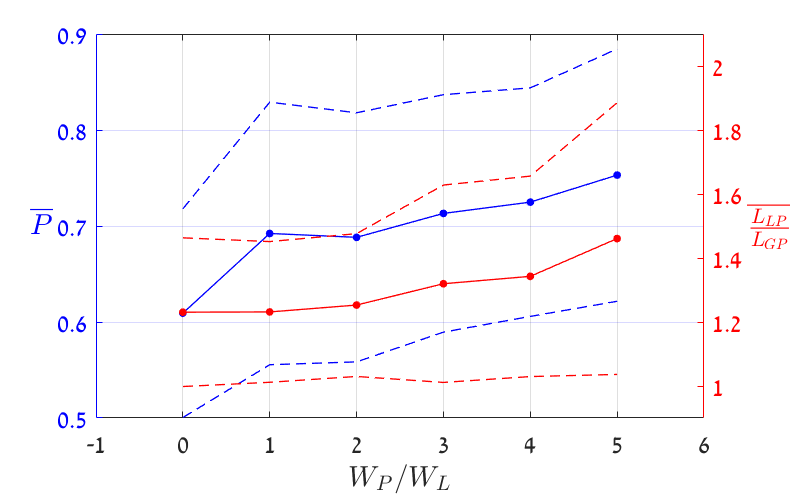}}
    \caption{LP-Dijkstra performance for different $W_P/W_L$ ratios using a binaric (a) and LPF (b) priority maps. The dashed lines are standard deviation margins.}
    \label{Dijkstra_Preformance_50_GPs}
\end{figure}

Table \ref{tab:Res1} summarize the results of trajectory planning statistics using different algorithms and in the case of surface sensing and binaric priority map (in the parentheses are standard deviations).  As expected, LP-Dijkstra gave the best results for average priority and total trajectory length, given the proper optimization weights.  However, it also had the slowest run-time. Different $W_P/W_L$ ratios were tested to determine what is a good operating point (Fig. \ref{Dijkstra_Preformance_50_GPs}). For small $W_P/W_L$ ratios short trajectories were generated with poor average priority values. For large $W_P/W_L$ ratios the algorithm focused on improving the average priority, even at the price of significantly increasing the average trajectory length and dispersion. A good operating point was found to be  $W_P/W_L=2$, improving average priority by $\sim$34\% to 0.82 while elongating the trajectory by only $\sim$7.2\% on average relative to the $W_P/W_L=0$ operating point. The shortest run-time was measured for WLP-$A^*$. GGWP generated trajectories only 0.8\% longer on average than LP-Dijkstra for $W_P=0$, and of similar average length compared to WLP-$A^*$ with $\sqrt{3}$ weighting factor. 

\begin{table} 
\centering 
\caption{Comparison of different trajectory planning algorithms for binaric priority map. Best results are marked in red.}
\begin{tabular}{p{0.095\textwidth}>{\centering}p{0.01\textwidth}>{\centering}p{0.01\textwidth}>{\centering}p{0.04\textwidth}>{\centering}p{0.04\textwidth}>{\centering\arraybackslash}p{0.09\textwidth}}

\toprule 
Algorithm & \textbf{$W_L$} & \textbf{$W_P$} &\textbf{$\overline{P}$} & \textbf{$\overline{\dfrac{L_{LP}}{L_{GP}}}$}  & Average Run- time [msec] \\ 
\midrule
 LP-Dijkstra  & 1&0& 0.61 \\ (0.11) & \textcolor{red}{1.25} \\ (0.23) &  20
 \\ %
  \hline
LP-Dijkstra  & 0.5&1 & \textcolor{red}{0.82} \\ (0.06) & 1.34 \\ (0.26) &  20
 \\ %
  \hline  
 LP-$A^*$   & 1&0& 0.62 \\(0.11) & \textcolor{red}{1.25} \\ (0.23) & 3.6\\ 
  \hline
  LP-$A^*$   & 0.5&1& 0.76 \\(0.09) & 1.37 \\ (0.28) & 3.6\\
  \hline
 WLP-$A^*$ ($W_{A^*}=\sqrt{3}$ )& 1&0& 0.61 \\(0.1) & 1.26 \\(0.25) & \textcolor{red}{2.5} \\ 
 \hline  %
 WLP-$A^*$ ($W_{A^*}=\sqrt{3}$ )& 0.5&1& 0.66 \\(0.12) & 1.4 \\(0.33) & \textcolor{red}{2.5} \\ 
 \hline  %
 GGWP  &  1&0& 0.59\\(0.12) & 1.26 \\(0.21) &  8.4\\

\bottomrule 
\end{tabular}
\label{tab:Res1} 
\end{table}

Table \ref{tab:Res_LPF} summarize the results when using a LPF priority map for the operating point $W_P/W_L=2$. In this case the differences in performance between the different algorithms are significantly reduced due to the spacial averaging nature of the low-pass filtering. Now in order to find a trajectory that has higher average priority, more significant spacial changes have to be made to the trajectory. These significant changes forces the trajectory to become longer and results in higher penalty. The priority improvement are therefore less favoured. 

\begin{table} 
\centering 
\caption{Comparison of different trajectory planning algorithms for LPF priority map.}
\begin{tabular}{p{0.095\textwidth}>{\centering}p{0.01\textwidth}>{\centering}p{0.01\textwidth}>{\centering}p{0.04\textwidth}>{\centering\arraybackslash}p{0.04\textwidth}}

\toprule 
Algorithm & \textbf{$W_L$} & \textbf{$W_P$} &\textbf{$\overline{P}$} & \textbf{$\overline{\dfrac{L_{LP}}{L_{GP}}}$}  \\
\midrule
 LP-Dijkstra  & 0.5&1 & 0.68\\ (0.12) &  1.28\newline (0.24) 
 \\ %
   \hline
LP-$A^*$   & 0.5&1& 0.65 \\(0.1) & 1.27 \newline (0.29)  \\
  \hline
 WLP-$A^*$ ($W_{A^*}=\sqrt{3}$ )& 0.5&1& 0.59 \\(0.12) & 1.26 \newline(0.21) \\ 
 
\bottomrule 
\end{tabular}
\label{tab:Res_LPF} 
\end{table}

\section{Conclusion and Future work}
The autonomous landing capability is important for increasing autonomy of UAVs. The landing mission requires a unique planning optimization process including a suitable local trajectory planning that can assist the mission while avoiding obstacle collision. The main contributions of this work is the introduction of a new optimization criterion and analysis methodology for trajectory planning, using a pre-planned global path and a priority map of landing site search regions on the ground. Algorithms were developed and evaluated using a simulation of real-life scenarios where only the surface of obstacles are detectable, and only at close enough distance. Trajectory planning using polynomial methods such as B\'ezier curves may fail in such cases of surface sensing. We  introduce and evaluate the performance of a weighted $A^*$ algorithm modified with the new optimization criterion (namely WLP-$A^*$). A greedy grid-based version of the cost wave propagation path planning algorithm (namely GGWP) was also evaluated and was shown to work well in practice for calculating short trajectories. Simulations show that in addition to finding feasible trajectories that avoid collisions with obstacles, our optimization process can significantly increase the probability of finding a proper landing site. 

However, in this work the UAV navigation data was assume to be perfect with no errors what so ever. The impact of navigation errors on obstacle mapping and consequently on trajectory planning may be profound, and therefore must also be included in future evaluations. The influence of other important parameters such as sensor detection range and mapping resolution must also be examined.

\printbibliography

\end{document}